\documentclass[sigconf]{acmart}
\AtBeginDocument{%
  }
\usepackage{multirow}
\usepackage{graphicx}
\usepackage{enumitem}
\usepackage{bm}
\usepackage{xcolor}
\usepackage{colortbl}
\usepackage{arydshln}
\usepackage{cleveref}
\usepackage{subcaption} 
\newcommand{\bestentry}[1]{\cellcolor{orange!20}\bfseries #1}
\newcommand{\secondentry}[1]{\cellcolor{gray!20}\underline{#1}}

\newenvironment{craddition}{\begingroup}{\endgroup}
\setlist[itemize]{leftmargin=*}

\copyrightyear{2026}
\acmYear{2026}
\setcopyright{cc}
\setcctype{by}
\acmConference[KDD '26]{Proceedings of the 32nd ACM SIGKDD Conference on Knowledge Discovery and Data Mining V.2}{August 09--13, 2026}{Jeju Island, Republic of Korea}
\acmBooktitle{Proceedings of the 32nd ACM SIGKDD Conference on Knowledge Discovery and Data Mining V.2 (KDD '26), August 09--13, 2026, Jeju Island, Republic of Korea}
\acmDOI{10.1145/3770855.3818905}
\acmISBN{979-8-4007-2259-2/2026/08}

\makeatletter
\newcount\gwpf@author@bx@count
\let\gwpf@typeset@author@bx\@typeset@author@bx
\def\@typeset@author@bx{%
  \advance\gwpf@author@bx@count by 1\relax
  \ifnum\gwpf@author@bx@count=5\relax
    \hspace*{-0.02\textwidth}%
    \author@bx@wd=.27\textwidth\relax
  \fi
  \ifnum\gwpf@author@bx@count=6\relax
    \author@bx@wd=.34\textwidth\relax
  \fi
  \ifnum\gwpf@author@bx@count=7\relax
    \author@bx@wd=.27\textwidth\relax
  \fi
  \gwpf@typeset@author@bx
}
\makeatother

\begin{document}

\title{Fourier Geometric Wind Power Forecasting with Numerical Weather Prediction}



\author{Shiyuan Piao}\authornote{Equal contribution.}
\affiliation{
  \institution{The Hong Kong University of Science and Technology}
    \city{Hong Kong}
  \country{China}
} 
\email{shiyuan.piao@connect.ust.hk}

\author{Zehui Fan}\authornotemark[1]
\affiliation{
  \department{Faculty of Engineering}
  \institution{
  The Chinese University of Hong Kong}
  \city{Hong Kong}
  \country{China}
}
\email{1155184696@link.cuhk.edu.hk}

\author{Yang Liu}\authornote{Corresponding author.}
\affiliation{
  \institution{The Chinese University of Hong Kong}
    \city{Hong Kong}
  \country{China}
}
\email{yangliu005@cuhk.edu.hk}

\author{Hong Cheng}
\affiliation{
  \institution{The Chinese University of Hong Kong}
    \city{Hong Kong}
  \country{China}
}
\email{hcheng@se.cuhk.edu.hk}

\author{Juepeng Zheng}
\affiliation{
  \institution{Sun Yat-Sen University}
    \city{Guangzhou}
  \country{China}
}
\email{zhengjp8@mail.sysu.edu.cn}

\author{Jie Zhou}
\affiliation{
  \institution{Tsinghua University}
  \institution{Goldwind Science and Technology Co.,Ltd}
      \city{Beijing}
  \country{China}
}
\email{j-zhou20@mails.tsinghua.edu.cn}

\author{Fugee Tsung}
\affiliation{
  \institution{The Hong Kong University of\\ Science and Technology}
    \city{Hong Kong}
  \country{China}
}
\email{season@ust.hk}
\renewcommand{\shortauthors}{Shiyuan Piao et al.}

\begin{abstract}
Accurate short-term wind power forecasting is essential for grid stability and operational planning, yet remains challenging due to the complex interactions between atmospheric conditions and turbine dynamics. However, existing methods fail to effectively incorporate weather forecasting with wind turbine data (i.e., SCADA), leading to suboptimal solutions.  To address this, we introduce a multimodal framework that integrates historical point-based SCADA data with grid-based Numerical Weather Prediction (NWP) forecasts, which is challenging due to heterogeneous input and the complex physical wind-turbine interactions. Our approach first explicitly decomposes inputs into scalar and vector features to better capture both site-specific and geometric dependencies and then incorporates a geometric encoder to extract rotation-invariant features from wind vectors. We further leverages a Fourier Neural Operator (FNO) architecture, which performs global convolutions in the frequency domain to efficiently model long-range spatiotemporal relationships. Extensive experiments on three real-world wind farms, with weather forecasting data, demonstrate that our model consistently outperforms state-of-the-art baselines, highlighting the effectiveness of its physically-informed design. The core implementation of our method is publicly available at: \url{https://github.com/shawn-sypiao/GWPF}.
\end{abstract}


\maketitle

\section{Introduction}

Global electricity consumption is entering a period of strong growth, as electrification expands in buildings, transport and industry, while additional demand is increasingly driven by cooling and data centres~\cite{ieaElectricity2025} . 
Wind power, being a clean and renewable energy source is considered a key low-emissions technology and a valid solution in this transition~\cite{ipccAR6WG3}. 
In particular, the IEA projects global wind generation to grow at about 11\% per year on average during 2025--2027, accounting for roughly one-third of the additional global electricity demand over the same period~\cite{ieaElectricity2025} . 
However, wind power output, i.e., SCADA data, is strongly conditioned on atmospheric variability and local flow conditions, which leads to pronounced uncertainty and temporal intermittency in electricity production~\cite{foley2012wind} . 
Such variability complicates system operation because power grids require continuous balancing between supply and demand, motivating accurate short-term wind power forecasting to improve dispatch, reserve scheduling and reliability~\cite{Haq2025Review}. 
Despite its practical benefits, wind power forecasting remains challenging due to the chaotic nature of atmospheric dynamics and the complex turbine--flow interactions within wind farms~\cite{foley2012wind}.

Existing approaches can be classified into physical and data-driven models. Physical models solve equations that incorporate Numerical Weather Prediction (NWP), turbine parameters, and farm geography to predict wind power. Nevertheless, their accuracy is limited by the oversimplified parametrization of complex physical processes and the high demands of computational resources. In contrast, data-driven models directly learn neural networks to approximate the wind power outputs, which can be broadly divided into two types: (1) \textbf{Spatio-temporal forecasting}. 
They integrate both spatial and temporal dependencies to enhance prediction accuracy by capturing how wind fields evolve over time and interact across geographical regions. A common architecture is Graph Neural Networks~\cite{DBLP:journals/fgcs/YuZLYGLYZY20,song2023gcn_wind}, where they typically represent turbines as nodes, construct edges among them via distance, and perform message passing thereon.

While these models are effective for exploiting cross-turbine correlations from SCADA histories, they are often deployed in a SCADA-only setting and thus may miss upcoming synoptic/mesoscale changes that are better captured by NWP \cite{tuncar2024review}; (2) \textbf{Wind power scenario generation}. Instead of outputting a single point estimate, probabilistic methods aim to produce calibrated predictive distributions or multiple coherent future trajectories (scenarios), which are crucial for risk-aware decision making in markets and system operation.
A classical NWP-based approach is EMOS~\cite{gneiting2005emos}, which statistically calibrates ensemble forecasts by fitting a parametric predictive distribution to correct bias and dispersion.
More recently, diffusion-based generative models have been developed for probabilistic time-series forecasting by learning a denoising process that can sample diverse yet coherent trajectories~\cite{rasul2021timegrad}.
In the wind power domain, conditional latent diffusion models have been proposed to generate short-term wind power scenarios with improved sharpness and calibration under complex uncertainty~\cite{dong2024cldm_wind}.Although these NWP-driven probabilistic approaches can quantify uncertainty, many of them do not explicitly model turbine operational states and the non-stationary turbine--wind relationship, which reduces forecasting accuracy.

However, it is not straightforward to incorporate data-driven wind power forecasting with numerical weather prediction due to:
\begin{itemize}
    \item \textbf{Data heterogeneity}. Insufficient resolution in NWP grids prevents direct access to wind conditions at individual turbine locations, presenting a fundamental data heterogeneity challenge. The grid-point mismatch leads to information loss and increased modeling complexity. It obscures the precise local wind conditions that directly drive each turbine, forcing the model to infer these conditions from coarse surrounding data. 

    \item \textbf{Physical interaction}. Wind power generation is fundamentally dominated by wind speed magnitude and wind direction. The misalignment between wind direction and nacelle yaw directly influences effective inflow and thus power conversion efficiency. Therefore, it is necessary to capture the physical interactions between gridded wind fields and point-level inflow, as well as the directional effects introduced by turbine layout and flow dynamics.

\end{itemize}

\begin{figure}
    \centering
    \includegraphics[width=0.95\linewidth]{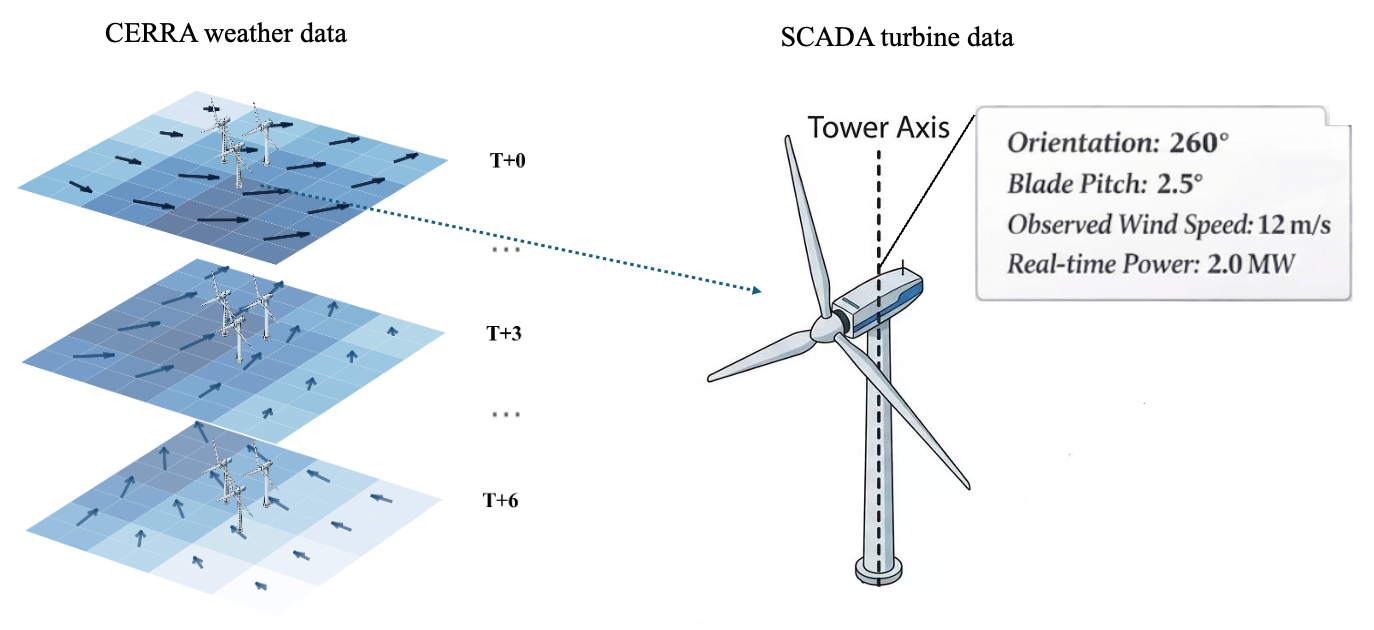}
    \caption{Data visualization. Gridded NWP weather forecasting provides the general wind field for site-specific SCADA turbine data.}
    \label{fig:intro}
    \vspace{-15pt}
\end{figure}

To address the above issue, we present a novel multimodal forecasting framework that integrates geometric priors and spectral feature fusion for short-term wind power prediction. We first employ a geometric encoder to explicitly model directional inductive biases by transforming raw wind vectors and nacelle angles into rotation-invariant scalar features,  improving its generalization across diverse wind regimes. Fourier Neural Operator (FNO) is flexible to adapt to varying resolutions.  We then employ an FNO-based Fusion to effectively combine the encoded turbine states with gridded weather forecasts by performing global convolutions in the frequency domain, thereby capturing complex, long-range spatiotemporal dependencies within and across modalities. Extensive experiments on three real-world wind farms and weather reanalysis data demonstrate that our framework consistently outperforms a comprehensive range of both spatio-temporal and wind power scenario generation models. Ablation studies further validate the contribution of each core component, confirming the necessity of both the geometric encoding and the neural operator fusion.

\section{Related Work}
\paragraph{\textbf{Spatio-temporal wind power forecasting}}

Deterministic wind power forecasting (WPF) has evolved into a sophisticated spatio-temporal point estimation task, shifting the focus from isolated time-series modeling to the integrated representation of regional meteorological systems~\cite{xu2025benchmark}. Within this paradigm, algorithms are trained to map historical SCADA-derived observations onto discrete future power sequences by minimizing distance-based loss functions~\cite{tuncar2024review}. 
The methodology for learning spatio-temporal dependencies is primarily centered on the following technical frameworks:
Spatial Interaction via Graph Neural Networks (GNNs)~\cite{DBLP:conf/kdd/0245Z0Z0025}: Recognizing that wind farms possess regional characteristics where turbines interact with one another through air movement, current research utilizes GNNs to model these complex dependencies~\cite{daenens2025stgnn_scada,qiu2024tsgnn_blockage,song2023gcn_wind,DBLP:journals/fgcs/YuZLYGLYZY20,DBLP:conf/iclr/LiuCZXZT0R24,zheng2025mesh,DBLP:conf/kdd/ZhengLL0R24}. By treating turbines or farms as nodes and their geographical or climatic relationships as edges, GNNs effectively capture regional interactions that single-site models fail to perceive, thereby enhancing the accuracy of the forecasting system.
Temporal Paradigms (AR vs. NAR): The temporal dimension is processed through either Autoregressive (AR) or Non-autoregressive (NAR) architectures. AR models, such as LSTM~\cite{hochreiter1997lstm} and GRU~\cite{cho2014gru}, iteratively propagate hidden states to preserve the continuity of historical trends, demonstrating high robustness in short-term horizons. Conversely, NAR models—including temporal-convolutional models such as TCN~\cite{bai2018tcn}, Transformer-based~\cite{zhou2021informer,DBLP:conf/iclr/LiuZCTZ0L25,wu2021autoformer}, and MLP-based~\cite{ekambaram2023tsmixer} architectures—generate the entire forecast sequence in a single pass. This generative approach leverages global dependencies more effectively, mitigates the accumulation of errors inherent in iterative methods, and increasingly matches or exceeds AR performance in long-term forecasting.

\paragraph{\textbf{Wind power scenario generation}}
In the current research landscape, wind power scenario generation is predominantly conducted within a probabilistic forecasting framework aimed at capturing the inherent stochasticity and meteorological volatility of wind energy to enhance grid reliability~\cite{bazionis2022prob_review,gneiting2005emos}. Methodologically, this task falls into three main streams: probability distribution estimation (e.g., DeepAR~\cite{salinas2020deepar}), which predicts parametric distribution parameters such as mean and variance via likelihood-based learning to enable trajectory sampling; quantile regression (e.g., QR-LSTM), a non-parametric approach that estimates conditional quantiles without assuming a specific distribution by minimizing the pinball loss across multiple quantile levels; and Bayesian neural networks (e.g., Bayesian-LSTM), which inject uncertainty directly into model weights through Bayesian inference—often approximated via variational methods or Monte Carlo dropout—to produce ensembles of diverse, risk-aware scenarios~\cite{gal2016dropout}. Architecturally, autoregressive models like LSTM and GRU are favored for their ability to preserve temporal dependencies and trends through iterative information propagation. Scenario quality is rigorously evaluated using a suite of metrics, including the Continuous Ranked Probability Score (CRPS) for overall distributional accuracy, and the Prediction Interval Coverage Probability (PICP) and Prediction Interval Normalized Average Width (PINAW) to assess the balance between interval reliability and sharpness~\cite{xu2025benchmark,gneiting2007scoring}.

Despite these advances, most existing methods still rely heavily on GNN and time series models that primarily use SCADA data and exhibit insufficient attention to spatiotemporal dependencies across wind farms. Meteorological data, such as numerical weather predictions, are typically incorporated through basic feature fusion approaches that do not adequately capture the complex spatial patterns, directional movements, or nonlinear evolution of weather systems~\cite{xu2025benchmark}. 
As a result, the generated scenarios often miss important atmospheric dynamics, revealing a clear need for models that better exploit the rich spatiotemporal structure in weather data.

\section{Background}
\subsection{Problem Definition}\label{sec:method_problem}
We formulate the short-term wind power forecasting task in a multi-modal setting that integrates historical turbine-level SCADA data and grid-based Numerical Weather Prediction (NWP) forecasts. The input features are decomposed into scalar and vector quantities to explicitly distinguish between geometric-agnostic and geometric variables. 
Let a wind farm consist of $N$ turbines. For each turbine $i$ at historical time step $t \in \{1, \dots, L\}$, we observe turbine scalar features $\bm{h}_{i}^{t}\in\mathbb{R}^{D_h}$ (e.g., wind speed magnitude, power output, operational status) and turbine vector features $\bm{q}_{i}^{t}\in\mathbb{R}^{D_q}$ (primarily the 2D wind vector). 
At the forecast anchor time $t = L$, the NWP system issues predictions for future lead times $\tau = 1, \dots, T$. For each $\tau$, it provides a gridded forecast over the farm region, including scalar fields $\bm{g}_{L+\tau}^{s}\in\mathbb{R}^{H\times W\times V_s}$ (e.g., temperature, pressure) and vector fields $\bm{g}_{L+\tau}^{v}\in\mathbb{R}^{H\times W\times V_v}$ (e.g., zonal and meridional wind components), where $H$ and $W$ denote the spatial grid dimensions, and $V_s$, $V_v$ are the numbers of scalar and vector variables, respectively.
The objective is to predict the hourly wind power output for all $N$ turbines over the next $T$ hours. Formally, we learn a mapping $f_\theta$:
\begin{equation}\label{eqn:task}
    \bm{y}_{L+1:L+T} = f_\theta\left( \{\bm{h}^{t}, \bm{q}^{t}\}_{t=1}^{L};\; \{\bm{g}_{L+\tau}^{s}, \bm{g}_{L+\tau}^{v}\}_{\tau=1}^{T} \right),
\end{equation}
where $\bm{y}_{L+1:L+T}\in\mathbb{R}^{N\times T}$ denotes the predicted power sequence.

\subsection{Fourier Neural Operator}
Fourier Transform is known to be an effective tool to extract features from periodic signals. Consider a sequence of $N$ grid-based real-valued observations of a function, denoted by $\bm{s} = (s_{1}, ..., s_{N}) \in \mathbb{R}^{N}$. 
The Discrete Fourier Transform (DFT), represented by $\mathcal{F}$, converts this sequence into the frequency domain with a periodicity of $2\pi$ as follows:
\begin{equation}\label{eqn:dft}
\begin{aligned}
    S_{k} = \sum_{n=1}^{N} s_{n} \cos \Bigl( 2\pi \frac{k}{N} n \Bigr) - i \sum_{n=1}^{N} s_{n} \sin \Bigl( 2\pi \frac{k}{N} n \Bigr) = A_{k} - B_{k} i,
\end{aligned}
\end{equation}
where $\bm{S} = \mathcal{F}(\bm{s}) = (S_{1}, ..., S_{N}) \in \mathbb{R}^{N}$ and $i$ is the imaginary unit. 
$A_{k}$ and $B_{k}$ are the real and imaginary parts of the complex number $S_{k}$ in the frequency domain, respectively. 
The inverse transformation, which reconstructs the original sequence from the frequency domain, is given by:
\begin{equation}
    s_{n} = \frac{1}{N} \sum_{k=1}^{N} S_{k} \Bigl( \cos \Bigl( 2\pi \frac{n}{N} k \Bigr) + i \sin \Bigl( 2\pi \frac{n}{N} k \Bigr) \Bigr),
\end{equation}
or equivalently by substituting $S_{k} = A_{k} - B_{k} i$ ,
\begin{equation}\label{eqn:idft}
     s_{n} = \frac{1}{N} \sum_{k=1}^{N} \Bigl( A_{k} \cos \Bigl( 2\pi \frac{n}{N} k \Bigr) - B_{k} \sin \Bigl( 2\pi \frac{n}{N} k \Bigr) \Bigr),
\end{equation}
where the imaginary unit is canceled out. We express the inverse transform as $\bm{s} = \mathcal{F}^{-1}(\bm{S})$ for symmetry.

The Fourier Neural Operator (FNO) is the neural network architecture designed to learn mappings between infinite-dimensional function spaces, making it particularly effective for modeling spatio-temporal physical systems. The core of FNO is to perform global convolutions in the Fourier frequency domain, allowing it to efficiently capture long-range spatial dependencies. The architecture stacks multiple Fourier layers, and each layer updates the hidden representation as follows:
\begin{equation}\label{eqn:fno-layer}
\begin{aligned}
    \bm{h}^{(l+1)} = \phi(\bm{h}^{(l)}\bm{W} + \mathcal{F}^{-1}(\bm{R}_\theta\odot\mathcal{F}(\bm{h}^{(l)})) + \bm{b}),
\end{aligned}
\end{equation}
where $\phi$ is a nonlinear activation function. $W, \bm{R}_\theta$ is learnable weight and complex-valued weight, respectively. $\bm{b}$ is a bias term.

\section{Method}
\begin{figure*}[!ht]
    \centering
    \includegraphics[width=\linewidth]{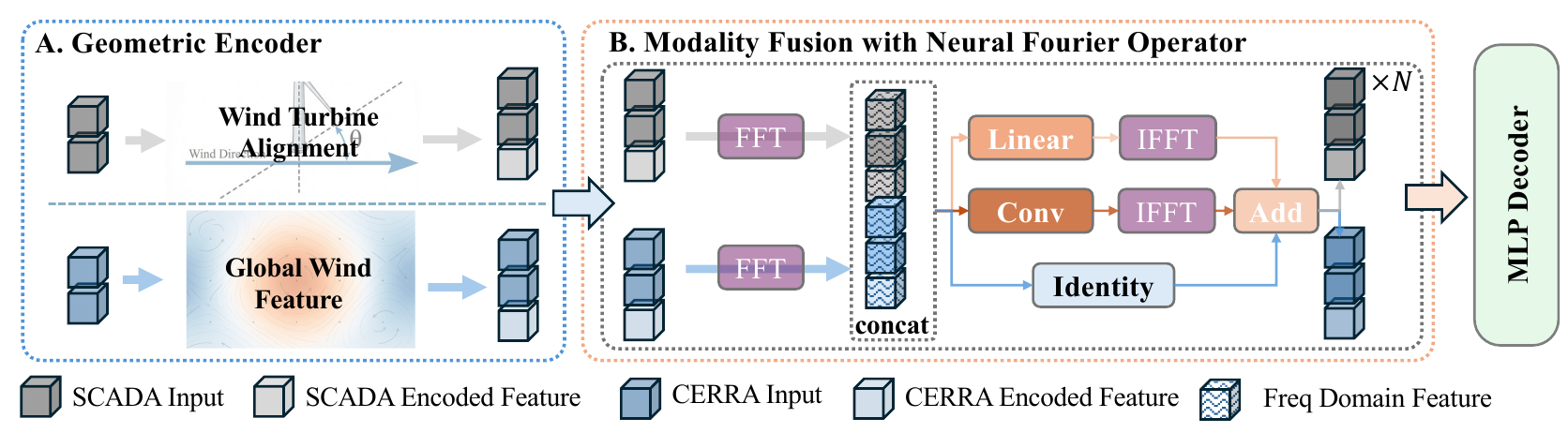}
    \caption{Overall framework of the proposed method.}
    \label{fig:framework}
\end{figure*}

In this section, we first elaborate on the overall design philosophy and introduce details of the proposed model as shown in \cref{fig:framework}. 

\subsection{Overall Framework}\label{sec:method_framework}
Our proposed framework consists of two core components: a geometric modeling module that encodes directional and spatial inductive biases, and a modality fusion module that effectively combines point-based turbine and grid-based weather forecasting embeddings. Specifically, the vectors are first transformed into a series of geometric-invariant features, which are defined as 
\begin{equation}\label{eqn:geo}
\begin{aligned}
    \{\bm{\hat{q}}^{t}\}_{t=1}^{L},\; \{\bm{\hat{g}}_{L+\tau}^{v}\}_{\tau=1}^{T} = \mu\left( \{\bm{q}^{t}\}_{t=1}^{L};\; \{\bm{g}_{L+\tau}^{v}\}_{\tau=1}^{T} \right),
\end{aligned}
\end{equation}
where $\mu$ is a non-parametric function. We concatenate them with the original scalar features and obtain the initial turbine and weather forecasting embeddings as follows:
\begin{equation}
\begin{aligned}
    \bm{h}_{s}^{(0)} = \sigma\left( \left[ \bm{h}^{t}, \bm{\hat{q}}^{t} \right]_{t=1}^{L} \right), \quad 
    \bm{h}_{g}^{(0)} = \sigma\left( \left[ \bm{g}_{L+\tau}^{s}, \bm{\hat{g}}_{L+\tau}^{v} \right]_{\tau=1}^{T} \right),
\end{aligned}
\end{equation}
where $\sigma$ denotes a Multilayer Perceptron (MLP). Then the embeddings are fed into multiple neural operator layers to learn the wind-turbine interactions. The $l$-th layer $\gamma_\theta^{(l)}$ updates the embeddings as follows:
\begin{equation}
\begin{aligned}
    \bm{h}_{s}^{(l+1)}, \bm{h}_{g}^{(l+1)}  = \gamma_\theta^{(l)}(\bm{h}_{s}^{(l)}, \bm{h}_{g}^{(l)}).
\end{aligned}
\end{equation}
Finally, a decoder is employed to predict the future hourly wind power output $\bm{\hat{y}}_{L+1:L+T}$. We minimize the discrepancy between the real and predicted values to optimize the model parameters:
\begin{equation}
\begin{aligned}
     \mathcal{L} = \frac{1}{N \times T} \left\| \bm{\hat{y}}_{L+1:L+T} - \bm{y}_{L+1:L+T} \right\|_F^2, 
\end{aligned}
\end{equation}
The geometric transformation function $\mu$ and the neural operator layers $\gamma$ are introduced in the following sections.

\subsection{Geometric Encoder}
Although raw wind vectors and nacelle angles can be directly fed into neural networks, such a representation ignores fundamental aerodynamic symmetries. Wind power generation is governed not by absolute directions, but by the \emph{relative orientation} between the incoming wind and the turbine’s nacelle. Similarly, large-scale atmospheric flow patterns, such as convergence zones or rotating eddies—strongly influence wake propagation and inter-turbine interactions, yet are invisible to models that only process raw grid-point wind components. To embed these physical priors, we replace raw directional inputs with deterministic, rotation-invariant scalar features via a non-parametric encoder $\mu$.

\paragraph{\textbf{\textit{SCADA inputs.}}}
For each turbine $i$ at time $t$, the SCADA system records active power $P_i^t$, wind speed $W_i^t$, wind direction $\theta_{\mathrm{dir},i}^t$, and nacelle yaw angle $\theta_{\mathrm{nac},i}^t$. To avoid discontinuities in angular representation, both directional quantities are encoded as sine--cosine pairs:
\[
(\sin\theta_{\mathrm{dir},i}^t, \cos\theta_{\mathrm{dir},i}^t), \quad (\sin\theta_{\mathrm{nac},i}^t, \cos\theta_{\mathrm{nac},i}^t).
\]
Together with $P_i^t$ and $W_i^t$, these form the base SCADA feature vector for turbine $i$:
\[
\mathbf{q}_i^t = \bigl( P_i^t,\; W_i^t,\; \sin\theta_{\mathrm{dir},i}^t,\; \cos\theta_{\mathrm{dir},i}^t,\; \sin\theta_{\mathrm{nac},i}^t,\; \cos\theta_{\mathrm{nac},i}^t \bigr)^\top .
\]

\paragraph{\textbf{\textit{NWP inputs.}}}
The NWP forecasting provides, for each of the $K = H \times W$ grid points covering the wind farm, zonal and meridional wind components $(u_{k,\tau}, v_{k,\tau})$ at future lead times $\tau = 1, \dots, T$ (corresponding to absolute time steps $L+\tau$), along with their fixed spatial coordinates $(x_k, y_k)$. The base NWP feature at grid point $k$ and lead time $\tau$ is thus:
\[
\mathbf{g}_{L+\tau,k}^v = (u_{k,\tau}, v_{k,\tau})^\top ,
\]
which captures only local wind vectors without explicit information about large-scale flow organization.

\begin{figure}[ht]
    \centering
    \begin{subfigure}[t]{0.48\linewidth}
        \centering
        \includegraphics[width=\linewidth]{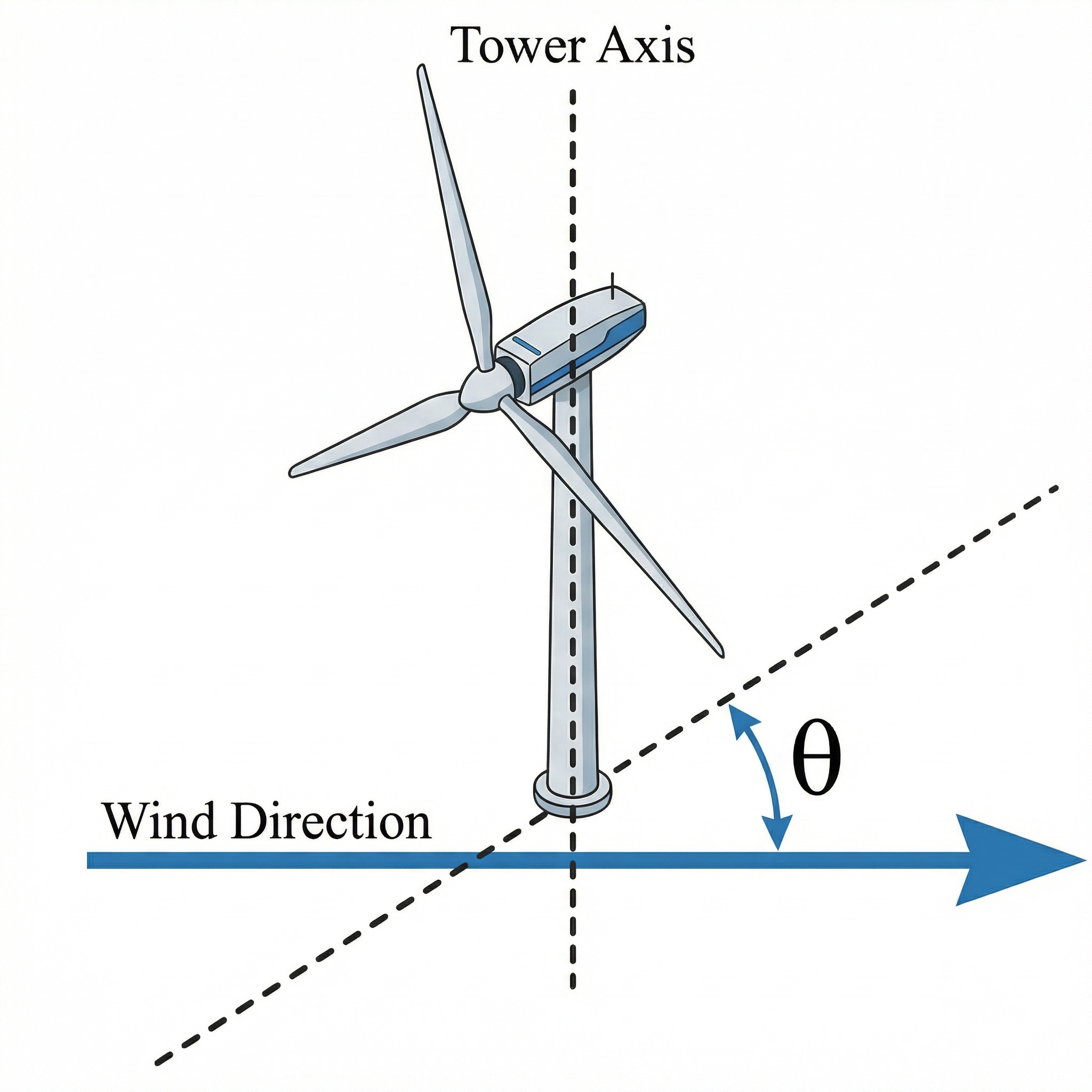}
        \caption{Wind-Turbine Alignment}
        \label{fig:method2-1}
    \end{subfigure}
    \hfill
    \begin{subfigure}[t]{0.48\linewidth}
        \centering
        \includegraphics[width=\linewidth]{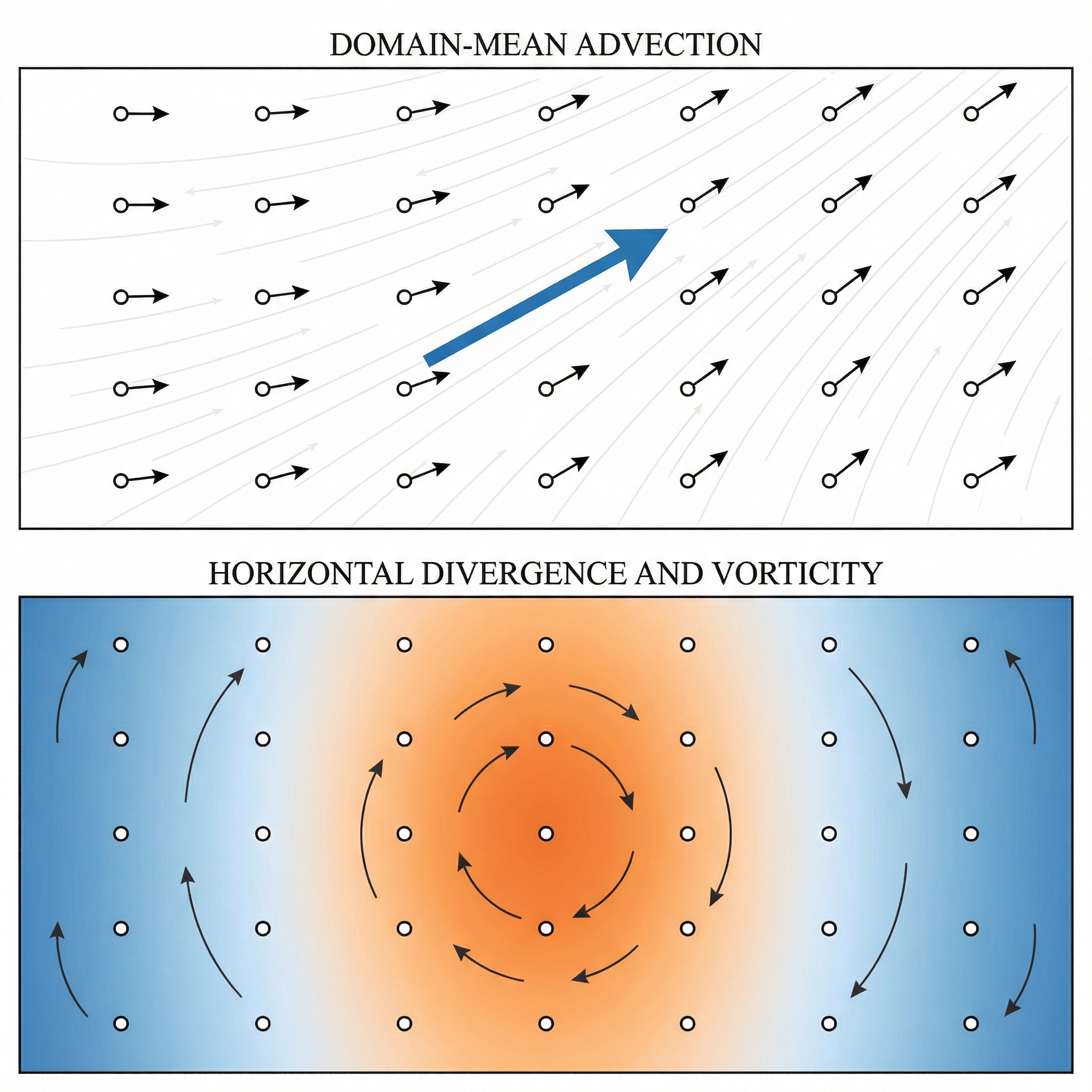}
        \caption{Global Wind Feature}
        \label{fig:method2-2}
    \end{subfigure}
    \caption{Illustration of the Geometric Encoder: (a) Wind Turbine Alignment; (b) Global Wind Feature.}
    \label{fig:placeholder}
\end{figure}

\paragraph{\textbf{\textit{Wind--turbine alignment.}}}
Turbine power output is governed by the relative orientation between incoming wind and nacelle yaw, not their absolute angles. For turbine $i$ at time $t$, let $\psi_{i,t} = \theta_{\mathrm{dir},i}^t - \theta_{\mathrm{nac},i}^t$ denote the yaw misalignment angle. The effective wind normal to the rotor plane scales with $\cos\psi_{i,t}$, which we compute continuously via trigonometric identities:
\begin{align}
\cos\psi_{i,t} &= \cos\theta_{\mathrm{dir},i}^t \cos\theta_{\mathrm{nac},i}^t + \sin\theta_{\mathrm{dir},i}^t \sin\theta_{\mathrm{nac},i}^t, \label{eq:cos_psi} \\
\sin\psi_{i,t} &= \sin\theta_{\mathrm{dir},i}^t \cos\theta_{\mathrm{nac},i}^t - \cos\theta_{\mathrm{dir},i}^t \sin\theta_{\mathrm{nac},i}^t, \label{eq:sin_psi}
\end{align}
as defined in Eqs.~\eqref{eq:cos_psi} and \eqref{eq:sin_psi}. We also include the power ramp rate $\Delta P_i^t = P_i^t - P_i^{t-1}$ to capture short-term dynamics. Combining these with the full nacelle orientation yields the derived SCADA feature for turbine $i$:
\begin{equation}
\hat{\mathbf{q}}_i^t = \bigl( \Delta P_i^t,\; \sin\theta_{\mathrm{nac},i}^t,\; \cos\theta_{\mathrm{nac},i}^t,\; \cos\psi_{i,t},\; \sin\psi_{i,t} \bigr)^\top, \label{eq:q_hat}
\end{equation}
defined in Eq.~\eqref{eq:q_hat}. This representation enforces rotational invariance: identical misalignment configurations produce identical features regardless of global heading.

\paragraph{\textbf{\textit{Global wind structure.}}}
While $\mathbf{g}_{L+\tau,k}^v$ describes local wind at grid point $k$ and lead time $\tau$, it omits coherent atmospheric structures that influence wake propagation across turbines. We therefore augment it with three physically meaningful descriptors:
\begin{itemize}
    \item \textbf{Horizontal divergence and vorticity}, estimated via least-squares linear regression over spatial coordinates at each lead time $\tau$:
\begin{align}
\begin{bmatrix} u_{k,\tau} \\ v_{k,\tau} \end{bmatrix} 
&\approx 
\begin{bmatrix} \beta_{u,0}^{(\tau)} \\ \beta_{v,0}^{(\tau)} \end{bmatrix} +
\begin{bmatrix} \beta_{u,1}^{(\tau)} & \beta_{u,2}^{(\tau)} \\ 
                \beta_{v,1}^{(\tau)} & \beta_{v,2}^{(\tau)} \end{bmatrix}
\begin{bmatrix} x_k \\ y_k \end{bmatrix}, \\
\mathrm{div}_\tau &= \beta_{u,1}^{(\tau)} + \beta_{v,2}^{(\tau)}, \quad
\mathrm{vort}_\tau = \beta_{v,1}^{(\tau)} - \beta_{u,2}^{(\tau)}.
\label{eq:div_vort}
\end{align}
as formulated in Eq.~\eqref{eq:div_vort}. Divergence quantifies flow convergence (affecting wake decay), while vorticity captures rotation (e.g., eddies or shear). If the spatial configuration is degenerate (e.g., $K < 3$), we set $\mathrm{div}_\tau = \mathrm{vort}_\tau = 0$.
\item \textbf{Domain-mean advection}, although not rotation-invariant, provides a general direction of entire fields. It is computed as the spatial average of the horizontal wind vector over all NWP grid points at lead time $\tau$:
\begin{equation}
    (\bar{u}_\tau, \bar{v}_\tau) = \frac{1}{K} \sum_{k=1}^{K} (u_{k,\tau}, v_{k,\tau}),
    \label{eq:domain_mean}
\end{equation}
which separates large-scale transport from local anomalies by removing the bulk motion component;
\end{itemize}
The resulting derived NWP feature at grid point $k$ and lead time $\tau$ is:
\begin{equation}
\hat{\mathbf{g}}_{L+\tau,k}^{v} = \bigl( u_{k,\tau},\; v_{k,\tau},\; \bar{u}_\tau,\; \bar{v}_\tau,\; \mathrm{div}_\tau,\; \mathrm{vort}_\tau \bigr)^\top, \label{eq:g_hat}
\end{equation}
defined in Eq.~\eqref{eq:g_hat}.
These derived features, $\hat{\mathbf{q}}_i^t$ for alignment and $\hat{\mathbf{g}}_{L+\tau,k}^{v}$ for flow structure, are concatenated with their respective base inputs ($\mathbf{q}_i^t$ and $\mathbf{g}_{L+\tau,k}^v$, plus any additional scalars) to form the complete input representations, which are subsequently projected into a shared embedding space by MLPs in the overall framework.

After constructing the physics-informed feature vectors $\hat{\mathbf{q}}_i^t$ and $\hat{\mathbf{g}}_{L+\tau,k}^{v}$, we form spatio-temporal input tensors for each modality:
\[
\mathbf{Q} \in \mathbb{R}^{L \times N \times d_s}, \quad
\mathbf{G} \in \mathbb{R}^{T \times K \times d_c},
\]
where $d_s$ and $d_c$ denote the total dimensions of the enriched SCADA and NWP features, respectively.
To enable cross-modal fusion, we first align the temporal and channel dimensions via two independent MLPs per modality. A temporal projection maps the sequence lengths $L$ and $T$ to a shared latent time dimension $d_{\mathrm{time}}$, while a channel projection maps $d_s$ and $d_c$ to a common hidden size $d_{\mathrm{hidden}}$. The initial embeddings are:
\[
\bm{h}_s^{(0)} \in \mathbb{R}^{d_{\mathrm{time}} \times N \times d_{\mathrm{hidden}}}, \quad
\bm{h}_g^{(0)} \in \mathbb{R}^{d_{\mathrm{time}} \times K \times d_{\mathrm{hidden}}}.
\]

\subsection{Modality Fusion}
Effectively integrating point-based SCADA measurements with low-resolution, grid-based NWP forecasts is challenging due to their mismatch in spatial support and temporal semantics. To bridge this gap, we process each modality through a frequency-domain transformation along the time axis, allowing their distinct spectral dynamics to be preserved and enhanced. The resulting time-domain representations are then concatenated across spatial nodes—turbines and grid points—to form a unified embedding that jointly captures local turbine behavior and regional weather patterns, thereby mitigating the granularity discrepancy while enabling long-range spatio-temporal modeling.

Formally, given SCADA embeddings $\bm{h}_s^{(l)} \in \mathbb{R}^{d_{\mathrm{time}} \times N \times d_{\mathrm{hidden}}}$ and NWP embeddings $\bm{h}_g^{(l)} \in \mathbb{R}^{d_{\mathrm{time}} \times K \times d_{\mathrm{hidden}}}$, we apply a shared frequency fusion module $\mathcal{M}$ to each:
\[
\tilde{\bm{h}}_s^{(l)} = \mathcal{M}(\bm{h}_s^{(l)}), \quad
\tilde{\bm{h}}_g^{(l)} = \mathcal{M}(\bm{h}_g^{(l)}).
\]
The updated representation is obtained by concatenating along the spatial dimension and applying a linear projection:
\begin{equation}\label{eqn:fuse}
    \bm{H}^{(l+1)} = \phi\left( [\tilde{\bm{h}}_s^{(l)}, \tilde{\bm{h}}_g^{(l)}] \bm{W} + \bm{b} \right) \in \mathbb{R}^{d_{\mathrm{time}} \times (N + K) \times d_{\mathrm{hidden}}},
\end{equation}
where $\bm{W} \in \mathbb{R}^{d_{\mathrm{hidden}} \times d_{\mathrm{hidden}}}$ and $\bm{b} \in \mathbb{R}^{d_{\mathrm{hidden}}}$ are learnable parameters applied independently to each time step and spatial location, and $\phi$ denotes a nonlinear activation (e.g., GELU).

\paragraph{\textbf{\textit{Frequency fusion} $\mathcal{M}$.}}
For an input $\bm{h} \in \mathbb{R}^{d_{\mathrm{time}} \times M \times d_{\mathrm{hidden}}}$ (with $M = N$ or $K$), we apply the Discrete Fourier Transform (DFT) along the time axis, yielding a complex-valued spectrum $\bm{S} = \mathcal{F}(\bm{h})$. We decompose it into real and imaginary parts $\bm{A} = \operatorname{Re}(\bm{S})$ and $\bm{B} = \operatorname{Im}(\bm{S})$, both in $\mathbb{R}^{d_{\mathrm{time}} \times M \times d_{\mathrm{hidden}}}$. Concatenating them along the channel dimension gives $[\bm{A}, \bm{B}] \in \mathbb{R}^{d_{\mathrm{time}} \times M \times 2 d_{\mathrm{hidden}}}$. We then apply a linear projection:
\[
\bm{Z} = [\bm{A}, \bm{B}] \bm{W}_{\mathrm{freq}}, \quad \bm{W}_{\mathrm{freq}} \in \mathbb{R}^{2 d_{\mathrm{hidden}} \times 2 d_{\mathrm{hidden}}},
\]
and split the result into enhanced real and imaginary parts:
\[
\bm{A}' = \bm{Z}_{:, :, :d_{\mathrm{hidden}}}, \quad \bm{B}' = \bm{Z}_{:, :, d_{\mathrm{hidden}}:}.
\]
An inverse DFT maps the enhanced spectrum back to the time domain:
\[
\tilde{\bm{h}} = \mathcal{F}^{-1}(\bm{A}' + i \bm{B}') \in \mathbb{R}^{d_{\mathrm{time}} \times M \times d_{\mathrm{hidden}}},
\]
which has the same shape as the input $\bm{h}$.
\begin{craddition}
This frequency fusion module follows the main ingredients of a Fourier Neural Operator layer: Fourier transform, learnable spectral mixing, inverse transform, and a local pathway with nonlinearity.
Here the spectral convolution is applied along the temporal axis rather than a spatial PDE grid, so the learnable spectral weights parameterize a global temporal kernel for multimodal forecasting.
\end{craddition}

\subsection{MLP Decoder}
The final fused embedding $\bm{H}^{(L)} \in \mathbb{R}^{B \times d_{\mathrm{time}} \times (N + K) \times d_{\mathrm{hidden}}}$ is flattened into a single vector:
\begin{equation}
    \bm{z} = \mathrm{flatten}\bigl( \bm{H}^{(L)} \bigr) \in \mathbb{R}^{ d_{\mathrm{time}} (N + K) d_{\mathrm{hidden}}}.
    \label{eq:decoder_input}
\end{equation}
This compact representation is then mapped to the forecast horizon via a two-layer MLP with ReLU activation.
\begin{equation}
    \bm{\hat{y}}_{L+1:L+H} = \mathrm{MLP}_{\mathrm{dec}}(\bm{z}) \in \mathbb{R}^{H \times N},
    \label{eq:decoder_output}
\end{equation}
as defined in Eq.~\eqref{eq:decoder_output}. No recurrent or attention mechanisms are used—the decoder relies entirely on the rich, physics-informed representation learned by the preceding modules.

\begin{table*}[!htbp]
    \setlength{\tabcolsep}{1.5mm}
    \centering 
    \caption{
        Overall performance comparison. 
        Best and second-best results (excluding \textit{Persistence} and \textit{Historical Mean}) 
        are highlighted in \textbf{bold with orange background} and underlined with gray background, respectively.
        Lower values are better ($\downarrow$).}
    \label{table:ai}
    \resizebox{\textwidth}{!}{
    \begin{tabular}{ccccccccccccccccccc}
    \toprule
    \multirow{3}{*}{\textbf{Model}}  
    & \multicolumn{4}{c}{\textbf{Kelmarsh Farm}} 
    & \multicolumn{4}{c}{\textbf{Penmanshiel Farm}} 
    & \multicolumn{4}{c}{\textbf{Hill of Towie Farm}} \\
    
    & \multicolumn{2}{c}{\textbf{RMSE $\downarrow$}} &  \multicolumn{2}{c}{\textbf{MAE $\downarrow$}} 
    & \multicolumn{2}{c}{\textbf{RMSE $\downarrow$}} &  \multicolumn{2}{c}{\textbf{MAE $\downarrow$}} 
    & \multicolumn{2}{c}{\textbf{RMSE $\downarrow$}} &  \multicolumn{2}{c}{\textbf{MAE $\downarrow$}} \\
    
    & $<3$ hour  & $<6$ hour & $<3$ hour  & $<6$ hour 
    & $<3$ hour  & $<6$ hour & $<3$ hour  & $<6$ hour 
    & $<3$ hour  & $<6$ hour & $<3$ hour  & $<6$ hour \\
    \midrule

    DLinear 
    &0.9044 &0.9266 &0.6729 &0.6992 
    &0.8823 &0.9002 &0.6474 &0.6690 
    &0.8041 &0.8374 &0.5857 &0.6101 \\
    
    TimeMixer 
    &1.3569 &1.3632 &1.0358 &1.0447 
    &1.2959 &1.3017 &1.0013 &1.0103 
    &1.2637 &1.2665 &0.9476 &0.9554 \\
    
    TimeXer 
    &1.3228 &1.3381 &1.0040 &1.0188 
    &1.2713 &1.2814 &0.9789 &0.9888 
    &1.2420 &1.2501 &0.9306 &0.9388 \\
    
    TSMixer 
    &0.8822 &0.9136 &0.6584 &0.6792 
    &0.8483 &0.8813 &0.6391 &0.6652 
    &0.7993 &0.8429 &0.5847 &0.6131 \\
    
    WPMixer 
    &1.3626 &1.3718 &1.0380 &1.0473 
    &1.2794 &1.2900 &0.9838 &0.9936 
    &1.2483 &1.2543 &0.9337 &0.9407 \\
    
    iTransformer 
    &1.3324 &1.3475 &1.0123 &1.0265 
    &1.2697 &1.2819 &0.9769 &0.9884 
    &1.2418 &1.2505 &0.9272 &0.9355 \\
    
    PatchTST 
    &1.3408 &1.3507 &1.0209 &1.0319 
    &1.2795 &1.2859 &0.9883 &0.9959 
    &1.2487 &1.2525 &0.9391 &0.9442 \\
    
    TimesNet 
    &1.2651 &1.2873 &0.9497 &0.9702 
    &1.2348 &1.2525 &0.9404 &0.9575 
    &1.2104 &1.2233 &0.9042 &0.9170 \\

    \hdashline
    
    TimeMoE 
    &1.4459 &1.4519 &1.1187 &1.1274 
    &1.3759 &1.3886 &1.0791 &1.0903 
    &1.3601 &1.3685 &1.0539 &1.0620 \\
    
    TiRex 
    &1.4659 &1.4533 &1.1480 &1.1349 
    &1.4070 &1.3945 &1.1099 &1.0991 
    &1.3750 &1.3637 &1.0720 &1.0630 \\
    
    Chronos-2
    &1.3902 &1.3954 &1.0604 &1.0695 
    &1.3546 &1.3536 &1.0432 &1.0484 
    &1.3443 &1.3380 &1.0248 &1.0245 \\
    
    \hdashline

     MPNN  
    &0.5685 &0.6679 &0.3797 &0.4513 
    &0.5664 &0.6747 &0.3650 &0.4419 
    &0.5903 &0.6764 &0.3780 &0.4412 \\
    
    D2STGNN  
    &\secondentry{0.5496} &0.6498 &0.3772 &0.4491 
    &0.5642 &0.6735 &0.3656 &0.4469 
    &0.5841 &0.6754 &0.3782 &0.4442 \\
    
    DCRNN  
    &0.5793 &0.6824 &0.3943 &0.4647 
    &0.6094 &0.7240 &0.4008 &0.4796 
    &0.6141 &0.7127 &0.3999 &0.4674 \\
    
    STID  
    &0.5621 &0.6612 &0.3852 &0.4563 
    &0.5689 &0.6763 &0.3745 &0.4524 
    &0.5939 &0.6801 &0.3782 &0.4424 \\
    
    MTGNN 
    &0.5508 &\secondentry{0.6496} &\secondentry{0.3734} &\secondentry{0.4448} 
    &0.5644 &0.6742 &\secondentry{0.3592} &\secondentry{0.4384} 
    &\secondentry{0.5770} &\secondentry{0.6683} &\secondentry{0.3686} &0.4337 \\
    
    AGCRN 
    &0.5516 &0.6531 &0.3755 &0.4488 
    &0.5631 &0.6723 &0.3663 &0.4449 
    &0.6084 &0.6901 &0.3919 &0.4530 \\

    \hdashline
    
    MiTSformer 
    &0.5553 &0.6589 &0.3801 &0.4524 
    &0.5674 &0.6706 &0.3726 &0.4498 
    &0.6073 &0.6949 &0.3955 &0.4551 \\
    
    HimNet 
    &0.5589 &0.6634 &0.3752 &0.4490 
    &0.5653 &0.6717 &0.3603 &0.4397 
    &0.5867 &0.6724 &0.3701 &\secondentry{0.4335} \\
    
    T-Graphormer 
    &0.5517 &0.6517 &0.3793 &0.4513 
    &0.5665 &0.6727 &0.3656 &0.4435 
    &0.5916 &0.6786 &0.3780 &0.4420 \\
    
    PatchSTG 
    &0.5604 &0.6603 &0.3854 &0.4559 
    &0.5663 &0.6735 &0.3732 &0.4504 
    &0.5928 &0.6868 &0.3761 &0.4430 \\
    
    HiSTGNN
    &0.5525 &0.6519 &0.3783 &0.4488 
    &\secondentry{0.5610} &\secondentry{0.6677} &0.3625 &0.4424 
    &0.6275 &0.7070 &0.4044 &0.4677 \\
    
    \hdashline
    
    Ours 
    &\bestentry{0.4412} &\bestentry{0.4562} &\bestentry{0.3001} &\bestentry{0.3112} 
    &\bestentry{0.4438} &\bestentry{0.4589} &\bestentry{0.2892} &\bestentry{0.2997} 
    &\bestentry{0.5073} &\bestentry{0.5237} &\bestentry{0.3264} &\bestentry{0.3379} \\
    
    \bottomrule
    \end{tabular}
    }
    \vspace{-10pt}
\end{table*}

\section{Experiment}

\subsection{Settings}

\paragraph{\textbf{\textit{Dataset.}}} We consider three UK wind farms, namely \textit{Kelmarsh}~\cite{zenodo_kelmarsh_2022}, \textit{Penmanshiel}~\cite{zenodo_penmanshiel_2022}, and \textit{Hill of Towie}~\cite{zenodo_hilloftowie_2025}, spanning calendar years 2016--2020.
For each farm, we collect (i) turbine-level SCADA measurements and (ii) gridded meteorological products from the CERRA~\cite{schimanke2021cerra_subdaily} dataset at 75 m height, including both reanalysis (analysis) and short-range forecasts. Additional details of these three datasets can be found in \cref{tab:dataset_summary}.
SCADA features include turbine power, observed wind speed, observed wind direction, nacelle position (yaw), and a curtailment/downtime indicator (\textit{Lost Production}).
To obtain reliable supervision, we mark an hourly target as valid only when (i) the power measurement is finite and (ii) \textit{Lost Production} equals zero; otherwise the target mask is set to 0 and ignored by loss/metrics.
We construct an input history at a step of 3 hours with length $L{=}9$ and a prediction horizon of $H{=}6$ at an hourly resolution.

\begin{table*}[t]
    \setlength{\tabcolsep}{1.2mm}
    \centering
    \caption{
        SCADA+CERRA fusion baselines: normalized MAE/RMSE on test set.
        All models use the same shared SCADA+NWP embedding; horizons 3\,h and 6\,h reported. 
        Best and second-best results are highlighted in \textbf{bold with orange background} and underlined with gray background, respectively.
        Lower values are better ($\downarrow$).
        }
    \label{table:nwpfusion-baselines}
    \resizebox{\textwidth}{!}{
    \begin{tabular}{lcccccccccccc}
    \toprule
    \multirow{3}{*}{\textbf{Model}} & \multicolumn{4}{c}{\textbf{Kelmarsh Farm}} & \multicolumn{4}{c}{\textbf{Penmanshiel Farm}} & \multicolumn{4}{c}{\textbf{Hill of Towie Farm}} \\
    \cmidrule(lr){2-5} \cmidrule(lr){6-9} \cmidrule(lr){10-13}
    & \multicolumn{2}{c}{\textbf{RMSE $\downarrow$}} & \multicolumn{2}{c}{\textbf{MAE $\downarrow$}} & \multicolumn{2}{c}{\textbf{RMSE $\downarrow$}} & \multicolumn{2}{c}{\textbf{MAE $\downarrow$}} & \multicolumn{2}{c}{\textbf{RMSE $\downarrow$}} & \multicolumn{2}{c}{\textbf{MAE $\downarrow$}} \\
    & $<3$ hour  & $<6$ hour & $<3$ hour  & $<6$ hour 
    & $<3$ hour  & $<6$ hour & $<3$ hour  & $<6$ hour 
    & $<3$ hour  & $<6$ hour & $<3$ hour  & $<6$ hour \\
    \midrule
    LSTM       & 0.5220 & 0.6154 & 0.3626 & 0.4312 & 0.5378 & 0.6388 & \secondentry{0.3499} & \secondentry{0.4245} & 0.5424 & 0.6227 & 0.3585 & 0.4185 \\
    MLP        & 0.5237 & 0.6145 & 0.3670 & 0.4363 & 0.5302 & 0.6300 & 0.3577 & 0.4314 & 0.5384 & \secondentry{0.6167} & 0.3585 & 0.4191 \\
    MPNN       & 0.5240 & 0.6180 & 0.3627 & 0.4361 & 0.5417 & 0.6451 & 0.3511 & 0.4254 & \secondentry{0.5372} & 0.6199 & \secondentry{0.3503} & \secondentry{0.4086} \\
    AGCRN      & \secondentry{0.5214} & \secondentry{0.6116} & 0.3739 & 0.4397 & \secondentry{0.5247} & \secondentry{0.6257} & 0.3585 & 0.4354 & 0.5448 & 0.6231 & 0.3665 & 0.4266 \\
    DLinear    & 0.5291 & 0.6254 & 0.3766 & 0.4473 & 0.5344 & 0.6362 & 0.3680 & 0.4449 & 0.5497 & 0.6293 & 0.3728 & 0.4332 \\
    iTransformer & 0.5250 & 0.6176 & \secondentry{0.3604} & \secondentry{0.4291} & 0.5357 & 0.6360 & 0.3530 & 0.4293 & 0.5436 & 0.6231 & 0.3544 & 0.4149 \\
    MTGNN & 0.5216 & 0.6140 & 0.3698 & 0.4396 & 0.5277 & 0.6292 & 0.3632 & 0.4403 & 0.5637 & 0.6388 & 0.3836 & 0.4423 \\
    HimNet & 0.5235 & 0.6173 & 0.3717 & 0.4414 & 0.5260 & 0.6272 & 0.3573 & 0.4354 & 0.5395 & 0.6195 & 0.3572 & 0.4186 \\
    \hdashline
    Ours       & \bestentry{0.4412} & \bestentry{0.4562} & \bestentry{0.3001} & \bestentry{0.3112} & \bestentry{0.4438} & \bestentry{0.4589} & \bestentry{0.2892} & \bestentry{0.2997} & \bestentry{0.5073} & \bestentry{0.5237} & \bestentry{0.3264} & \bestentry{0.3379} \\
    \bottomrule
    \end{tabular}
    }
    \vspace{-10pt}
\end{table*}

\paragraph{\textbf{\textit{Baselines.}}}
We compare against a comprehensive set of baselines summarized in \cref{table:ai,table:nwpfusion-baselines,tab:nwp-only-baselines}.
They are grouped as follows:
\begin{itemize}
  \item \textbf{Time-series models:} LSTM~\cite{hochreiter1997lstm}, DLinear~\cite{zeng2023dlinear}, TimeMixer~\cite{wang2024timemixer}, TimeXer~\cite{wang2024timexer}, TSMixer~\cite{ekambaram2023tsmixer}, WPMixer~\cite{murad2025wpmixer}, iTransformer~\cite{liu2024itransformer}, PatchTST~\cite{nie2023patchtst}, and TimesNet~\cite{wu2023timesnet}.
  \item \textbf{Foundation time-series models:} TimeMoE~\cite{shi2025timemoe}, TiRex~\cite{auer2025tirex}, and Chronos-2~\cite{ansari2025chronos2}.
  \item \textbf{Spatio-temporal models:} DCRNN~\cite{li2018dcrnn}, MTGNN~\cite{wu2020mtgnn}, AGCRN~\cite{bai2020agcrn}, STID~\cite{shao2022stid}, D2STGNN~\cite{shao2022d2stgnn}, MPNN~\cite{gilmer2017mpnn}, MiTSformer~\cite{chen2024mitsformer}, HimNet~\cite{dong2024himnet}, T-Graphormer~\cite{bai2025tgraphormer}, PatchSTG~\cite{fang2025patchstg}, and HiSTGNN~\cite{ma2023histgnn}.
  \item \textbf{Scenario generation models:} CLDM~\cite{dong2024cldm_wind}, Basic Diffusion~\cite{ho2020ddpm}, WiDLinear~\cite{ZhangWeatherinformedProbabilisticForecasting2025}, and WiTFT~\cite{ZhangWeatherinformedProbabilisticForecasting2025}.
\end{itemize}
All baselines are evaluated under the same train/val/test split, the same input/output definition $(L,H)$, and the same masking rule to ensure fair comparison.
For probabilistic methods, we report deterministic errors using the predictive mean/median (consistent across methods) when computing MAE/RMSE.
The implementation details can be found in the Appendix.


\subsection{Overall Comparison}
\paragraph{\textbf{\textit{SCADA-only baselines.}}}
\cref{table:ai} and \cref{table:nwpfusion-baselines} provide the main experimental results of the baselines with only SCADA input and the baselines with SCADA+CERRA fusion input. We report masked MAE and RMSE on two horizon ranges: <3h (steps 1–3) and <6h (steps 1–6). MAE summarizes typical errors, while RMSE is more sensitive to large deviations. We observe that our model achieved the best performance compared to all the baselines in all datasets. In the following, we discuss the results of these two tables, respectively.

All baselines in table 2 are only input the scada data and divided into 4 groups according to our summary in the experimental setting. Overall, our proposed model, which combines SCADA and CERRA data, achieves the best performance compared to all other baselines on all three wind farms and both horizons, consistently reducing both MAE and RMSE.
A key observation is that SCADA-only models exhibit a clear performance ceiling, especially at the longer horizon. Compared with the strongest SCADA-only baseline (e.g., MTGNN and HiSTGNN) in each setting, our method yields \textbf{about 19--21\% }lower errors at <3h on Kelmarsh and Penmanshiel, and \textbf{about 30--32\% }lower errors at <6h; on the larger Hill of Towie farm, the gains remain consistent at \textbf{about 11--12\% }(<3h) and \textbf{about 22\% }(<6h). The larger relative improvements at <6h align with the limitation of SCADA-only forecasting: as the horizon increases, turbine measurements become less informative about upcoming mesoscale changes, while drift accumulates in purely extrapolative predictions.

Generic time-series forecasters and large pre-trained time-series models consistently show the weakest performance. A plausible explanation is that they mainly capture temporal regularities while treating the farm as a flat multivariate sequence, lacking explicit spatial inductive bias for cross-turbine dependence, which leads to the larger error growth as the forecast horizon extends.
In contrast, spatio-temporal models are generally stronger, suggesting that explicitly modeling turbine-to-turbine interactions is beneficial, and this benefit becomes more evident on larger farms where spatial structure is richer. The best-performing ST variants likely benefit from more expressive spatial aggregation and temporal modeling (e.g., adaptive dependency learning), but still fall short of our proposed model, indicating that SCADA-only learning cannot fully resolve longer-horizon uncertainty without future atmospheric information.
These results motivate the hybrid input comparison in \cref{table:nwpfusion-baselines} While incorporating CERRA forecasts has the potential to break the SCADA-only ceiling, it is non-trivial due to data heterogeneity and physical interaction, which are the problems our model trying to solve.

\subsection{NWP Baselines Comparison}
\begin{table}[t]
\centering
\caption{NWP-only baselines: normalized MAE at $<6$ hour horizon by dataset.}
\label{tab:nwp-only-baselines}
\resizebox{\linewidth}{!}{
\begin{tabular}{l|ccccc}
\toprule
\textbf{Method} & \textbf{Kelmarsh Farm} & \textbf{Penmanshiel Farm} & \textbf{Hill of Towie Farm} & \textbf{Average} \\
\midrule
CLDM     & 0.3278 & \secondentry{0.3202} & 0.3618 & 0.3366 \\
Basic Diffusion & 0.3934 & 0.3842 & 0.4342 & 0.4039 \\
WiDLinear & 0.4173 & 0.4141 & 0.4376 & 0.4230 \\
WiTFT    & \secondentry{0.3238} & 0.3227 & \secondentry{0.3559} & \secondentry{0.3341} \\
\hdashline
Ours     & \bestentry{0.3112} & \bestentry{0.2997} & \bestentry{0.3379} & \bestentry{0.3163} \\
\bottomrule
\end{tabular}
}
\vspace{-10pt}
\end{table}
\paragraph{\textbf{\textit{NWP-only baselines.}}}
As shown in \cref{tab:nwp-only-baselines},
compared to NWP-only baselines using only CERRA inputs, our method achieves the lowest average normalized MAE across all three wind farms. Specifically, it yields relative improvements of approximately 5.9\% over CLDM, 25.2\% over WiDLinear, and 5.3\% over WiTFT.
The advantage over NWP-only baselines indicates that future weather forecasts alone are not sufficient for accurate turbine-level power prediction.
Although CERRA provides future atmospheric information, it does not directly encode the current turbine operating state, such as recent power output, local wind speed, and nacelle orientation.
By combining CERRA forecasts with SCADA histories and geometry-aware wind--turbine representations, our model better captures the non-stationary mapping from forecasted wind conditions to actual power generation.

\begin{table*}[ht]
\setlength{\tabcolsep}{1.2mm} 
\centering
\caption{Ablation study (main): overall performance. 
Each entry is formatted as ``value\textsubscript{+$\Delta$}'', 
where value is the metric rounded to three decimal places, 
and $\Delta$ denotes the increase in error relative to the full model (``Ours''). 
Lower RMSE and MAE are better.}
\label{tab:ablation-main}
\resizebox{\textwidth}{!}{
\begin{tabular}{lcccccccccccc}
\toprule
\multirow{3}{*}{\textbf{Method}} 
& \multicolumn{4}{c}{\textbf{Kelmarsh Farm}} 
& \multicolumn{4}{c}{\textbf{Penmanshiel Farm}} 
& \multicolumn{4}{c}{\textbf{Hill of Towie Farm}} \\
\cmidrule(lr){2-5} \cmidrule(lr){6-9} \cmidrule(lr){10-13}
& \multicolumn{2}{c}{\textbf{RMSE $\downarrow$}} 
& \multicolumn{2}{c}{\textbf{MAE $\downarrow$}} 
& \multicolumn{2}{c}{\textbf{RMSE $\downarrow$}} 
& \multicolumn{2}{c}{\textbf{MAE $\downarrow$}} 
& \multicolumn{2}{c}{\textbf{RMSE $\downarrow$}} 
& \multicolumn{2}{c}{\textbf{MAE $\downarrow$}} \\
& $<3$h & $<6$h & $<3$h & $<6$h 
& $<3$h & $<6$h & $<3$h & $<6$h 
& $<3$h & $<6$h & $<3$h & $<6$h \\
\midrule
w/o FFT 
& 0.471\textsubscript{+0.030} & 0.475\textsubscript{+0.018} & 0.322\textsubscript{+0.022} & 0.327\textsubscript{+0.016}
& 0.481\textsubscript{+0.037} & 0.482\textsubscript{+0.023} & 0.315\textsubscript{+0.026} & 0.319\textsubscript{+0.019}
& 0.530\textsubscript{+0.023} & 0.540\textsubscript{+0.016} & 0.342\textsubscript{+0.016} & 0.350\textsubscript{+0.012} \\

\textbf{w/o NFL} 
& \textbf{0.555\textsubscript{+0.114}} & \textbf{0.664\textsubscript{+0.208}} & \textbf{0.379\textsubscript{+0.078}} & \textbf{0.455\textsubscript{+0.144}}
& \textbf{0.560\textsubscript{+0.116}} & \textbf{0.669\textsubscript{+0.210}} & \textbf{0.366\textsubscript{+0.066}} & \textbf{0.449\textsubscript{+0.149}}
& \textbf{0.594\textsubscript{+0.087}} & \textbf{0.677\textsubscript{+0.153}} & \textbf{0.385\textsubscript{+0.058}} & \textbf{0.445\textsubscript{+0.107}} \\

\textbf{w/o CERRA} 
& \textbf{0.550\textsubscript{+0.109}} & \textbf{0.658\textsubscript{+0.202}} & \textbf{0.377\textsubscript{+0.077}} & \textbf{0.453\textsubscript{+0.142}}
& \textbf{0.564\textsubscript{+0.120}} & \textbf{0.670\textsubscript{+0.211}} & \textbf{0.366\textsubscript{+0.066}} & \textbf{0.447\textsubscript{+0.147}}
& \textbf{0.594\textsubscript{+0.087}} & \textbf{0.679\textsubscript{+0.155}} & \textbf{0.388\textsubscript{+0.062}} & \textbf{0.448\textsubscript{+0.110}} \\

w/o Physical Alignment
& 0.447\textsubscript{+0.006} & 0.466\textsubscript{+0.010} & 0.304\textsubscript{+0.003} & 0.316\textsubscript{+0.005}
& 0.455\textsubscript{+0.011} & 0.468\textsubscript{+0.010} & 0.289\textsubscript{+0.000} & 0.303\textsubscript{+0.003}
& 0.511\textsubscript{+0.004} & 0.533\textsubscript{+0.009} & 0.329\textsubscript{+0.002} & 0.341\textsubscript{+0.003} \\

w/o Global Wind 
& 0.460\textsubscript{+0.019} & 0.477\textsubscript{+0.021} & 0.312\textsubscript{+0.012} & 0.328\textsubscript{+0.017}
& 0.469\textsubscript{+0.025} & 0.490\textsubscript{+0.031} & 0.303\textsubscript{+0.014} & 0.320\textsubscript{+0.021}
& 0.535\textsubscript{+0.028} & 0.556\textsubscript{+0.032} & 0.345\textsubscript{+0.018} & 0.358\textsubscript{+0.020} \\
\hdashline
Full Structure
& 0.441 & 0.456 & 0.300 & 0.311 
& 0.444 & 0.459 & 0.289 & 0.300 
& 0.507 & 0.524 & 0.326 & 0.338 \\
\bottomrule
\end{tabular}
}
\vspace{-10pt}
\end{table*}

\setlength{\textfloatsep}{8pt plus 2pt minus 1pt}
\begin{table}[t]
\centering
\vspace{-4pt}
\caption{Feature-selection analysis with c-STG at 6h MAE. The reference full-model 6h MAE is 0.316.}
\label{tab:cr-prior-completeness}
\small
\setlength{\tabcolsep}{3pt}
\resizebox{0.94\linewidth}{!}{%
\begin{tabular}{llcc}
\toprule
\textbf{Group} & \textbf{Feature} & \textbf{Gate} & \textbf{MAE w/o} \\
\midrule
Original & physical alignment & 0.281 & 0.320 (+0.004) \\
Original & global-wind vorticity & 0.224 & 0.323 (+0.007) \\
Original & global-wind divergence & 0.169 & 0.322 (+0.006) \\
New & wake tendency & 0.141 & 0.317 (+0.001) \\
New & spatial NWP speed anomaly & 0.132 & 0.315 (-0.001) \\
New & turbulence intensity & 0.053 & 0.314 (-0.002) \\
\bottomrule
\end{tabular}
}
\end{table}

\paragraph{\textbf{\textit{Additional SCADA+CERRA fusion baselines.}}}
\cref{table:nwpfusion-baselines} evaluates NAR baselines under a shared SCADA+NWP embedding scheme. Concretely, the decoder input is formed by concatenating the historical tokens with the NWP block, and then applying the standard Informer-style embedding~\cite{zhou2021informer} as the sum of value, positional, and temporal embeddings. Compared with SCADA-only forecasting (\cref{table:ai}), adding CERRA consistently reduces both MAE and RMSE on all three wind farms, with larger gains at longer horizons. 
Within these baselines, models with explicit spatial inductive bias (e.g., MPNN/AGCRN) are generally strongest, while simpler/linear backbones(MLP/ARMLP and especially DLinear) lag behind, likely due to limited capacity to model heterogeneous inputs and nonlinear cross-modal interactions. Despite these improvements, our method remains best on all farms and horizons, delivering \textbf{about 15--17\%} lower errors at <3h and \textbf{25--29\% }at <6h on Kelmarsh and Penmanshiel, and \textbf{about 6--7\%} at <3h and \textbf{15--19\%} at <6h on Hill of Towie compared with the strongest fusion baseline.
This gap indicates that the performance bottleneck is not only the availability of NWP itself, but how SCADA and NWP are fused: simple token-level concatenation and generic embeddings are insufficient to capture vector-valued physical relations (e.g., direction-dependent inflow, yaw misalignment, and transport-driven linkage between turbine-point states and global wind fields). By explicitly encoding geometric invariances and performing a fusion mechanism tailored to this forecasting setting, our model better exploits NWP signals and yields larger gains, especially at longer horizons.

\begin{craddition}
\paragraph{\textbf{Operational interpretation of error reductions.}}
\enlargethispage{2\baselineskip}
A 0.01 reduction in normalized MAE corresponds to roughly 5.8--7.2 kW for each turbine-level forecasted step on the three farms, with the detailed conversion reported in \cref{tab:cr-kw-scale}. Thus the gains in \cref{table:nwpfusion-baselines,tab:ablation-main} accumulate over turbines, rolling forecast windows, and repeated dispatch intervals, reducing aggregate imbalance risk at the $<6$h horizon where operators can still update reserves, correct dispatch schedules, and adjust intraday bids. This is also the horizon at which NWP is most useful: the SCADA history remains reliable for immediate persistence, while future flow fields provide the lead-time information needed to distinguish sustained ramps from transient turbine-level fluctuations.
\end{craddition}

\vskip -8pt
\subsection{Empirical Analysis}
\vskip -3pt

\paragraph{\textbf{\textit{Ablation study on main blocks.}}}
We conduct an ablation study (\cref{tab:ablation-main}) to quantify the contribution of each key component in our model.
\begin{craddition}
The variants cover three functional blocks: future-weather input (\textbf{w/o CERRA}), spectral fusion (\textbf{w/o FFT}, \textbf{w/o NFL}), and geometric priors (\textbf{w/o Physical Alignment}, \textbf{w/o Global Wind}).
Overall, \cref{tab:ablation-main} shows that the full model performs best on all three farms, and every ablated variant introduces a measurable degradation.
The largest drops occur for w/o CERRA and w/o NFL, confirming that the gain depends on both future atmospheric information and an effective spectral fusion mechanism.
Removing CERRA reduces the task to a SCADA-only forecast and therefore loses direct information about upcoming weather evolution, which is especially harmful at the longer horizon.
Removing NFL keeps both SCADA and CERRA available but replaces the learnable spectral operator with a simple projection, indicating that the fusion block is essential rather than a minor architectural detail.
w/o FFT gives a smaller but consistent drop, suggesting that the frequency-domain representation helps align slowly evolving NWP signals with short-term power fluctuations.
For geometric priors, removing physical alignment causes a modest decline because raw wind direction, nacelle angle, and wind speed already contain partially redundant directional cues.
Removing global wind hurts more, showing that advection, divergence, and vorticity provide non-local farm-level flow context that cannot be fully recovered from independent grid-point wind vectors.

\paragraph{\textbf{\textit{Fine-grained geometric-prior analysis.}}}
To complement the coarse block-level ablation in \cref{tab:ablation-main}, \cref{tab:cr-prior-completeness} provides a fine-grained analysis of the geometric prior features.
We augment the original geometric features with turbulence intensity, wake tendency, and spatial NWP speed anomaly, and then apply c-STG~\cite{sristi2024cstg} with the backbone and training protocol fixed.
\cref{tab:cr-prior-completeness} shows that the original physical alignment and global-wind features receive larger gate mass and cause larger leave-one-feature-out degradation; parentheses report the MAE change relative to the full model.
This feature-selection result is consistent with \cref{tab:ablation-main}: the geometric-prior block is useful because it encodes physically meaningful turbine--flow relations, rather than merely adding extra weather-derived covariates.
Together, these two analyses attribute the gains to future weather inputs, spectral fusion, and geometry-aware priors.
\end{craddition}

\begin{craddition}
\paragraph{\textbf{\textit{Operational case study.}}}
The ablation and fine-grained results in \cref{tab:ablation-main,tab:cr-prior-completeness} suggest how the model behaves under a representative steady high-wind regime, where the background wind speed is strong but individual turbines can still depart from the ideal power curve.
In this regime, yaw misalignment is the first explanatory signal: even when the NWP wind speed indicates favorable inflow, a large wind--nacelle angle reduces the effective rotor-facing component and explains turbine-specific under-generation.
The global-wind features then account for residual farm-level structure.
Divergence indicates whether the local flow is spreading or converging, and vorticity captures rotating or sheared flow that can alter wake transport across neighboring turbines.
These signals are activated jointly: physical alignment corrects the local turbine response, while divergence and vorticity modulate how the correction propagates through the farm, which explains why removing global wind is more harmful at the longer $<6$h horizon.
\end{craddition}

\paragraph{\textbf{\textit{Sensitivity to the number of NFL layers.}}}
\cref{tab:ablation-nfl-layers} reports the impact of varying the number of NFL layers (with FFT) on forecasting accuracy. The model performs worst when no NFL layer is used ($n_{\text{NFL}}{=}0$), indicating that neural fourier layer is necessary for effective fusion.
The optimal NFL depth differs across datasets. Kelmarsh (N=6) achieves the best result at $n_{\text{NFL}}{=}1$ (3h/6h MAE: 0.2997/0.3116). Penmanshiel (N=14) performs best at $n_{\text{NFL}}{=}2$ (0.2873/0.2999), while Hill of Towie (N=21) reaches its optimum at $n_{\text{NFL}}{=}3$ (0.3275/0.3385). This trend suggests that as the dataset scale (number of turbines) increases, using a deeper NFL improves performance by providing greater spectral modeling capacity. However, further increasing the depth beyond the optimal setting leads to marginal degradation, showing that overly deep NFL stacks may overfit and hinder generalization.

\begin{table}[H]
\small
\centering
\caption{Comparison on number of NFL layers (w/ FFT). 3h / 6h MAE.}
\label{tab:ablation-nfl-layers}
\resizebox{0.94\linewidth}{!}{%
\begin{tabular}{lccc}
\toprule
NFL & Kelmarsh(N=6) & Penmanshiel(N=14) & Hill of Towie(N=21) \\
\midrule
0  & 0.3779 / 0.4538 & 0.3652 / 0.4444 & 0.3830 / 0.4460 \\
1       & \textbf{0.2997 / 0.3116} & 0.2899 / 0.3020 & 0.3299 / 0.3409 \\
2       & 0.3007 / 0.3126 & \textbf{0.2873 / 0.2999} & 0.3287 / 0.3414 \\
3       & 0.3008 / 0.3121  & 0.2899 / 0.3014 & \textbf{0.3275 / 0.3385}  \\
4       & 0.3011 / 0.3116 & 0.2921 / 0.3032 & 0.3294 / 0.3408  \\
\bottomrule
\end{tabular}
}
\vspace{-3pt}
\parbox{0.94\linewidth}{\footnotesize w/o NFL: $n_{\mathrm{NFL}}=0$ (no NFL layers). Cell format: 3h MAE / 6h MAE.}
\end{table}
\vskip -10pt

\begin{craddition}
\paragraph{\textbf{Decoder variants.}}
We further vary only the decoder while keeping the encoder and fusion module fixed, and report mean normalized test MAE over Kelmarsh, Penmanshiel, and Hill of Towie.
\cref{tab:cr-decoder-variants} shows that larger sequence decoders provide only marginal gains over the MLP decoder, so the MLP is an efficient default for the proposed architecture.

\begin{table}[H]
\centering
\caption{Decoder variants. Metrics are mean normalized test MAE over the three UK farms.}
\label{tab:cr-decoder-variants}
\resizebox{0.94\linewidth}{!}{%
\begin{tabular}{lccc}
\toprule
\textbf{Decoder} & \textbf{\#Params} & \textbf{$<3$h MAE} & \textbf{$<6$h MAE} \\
\midrule
MLP & 0.12M & $0.304 \pm 0.017$ & $0.316 \pm 0.016$ \\
LSTM & 0.25M & $0.302 \pm 0.019$ & $0.314 \pm 0.017$ \\
iTransformer & 0.32M & $0.301 \pm 0.017$ & $0.313 \pm 0.016$ \\
\bottomrule
\end{tabular}
}
\end{table}
\end{craddition}

\paragraph{\textbf{\textit{Model size.}}}
\begin{figure}[H]
    \centering
    \includegraphics[width=0.99\linewidth]{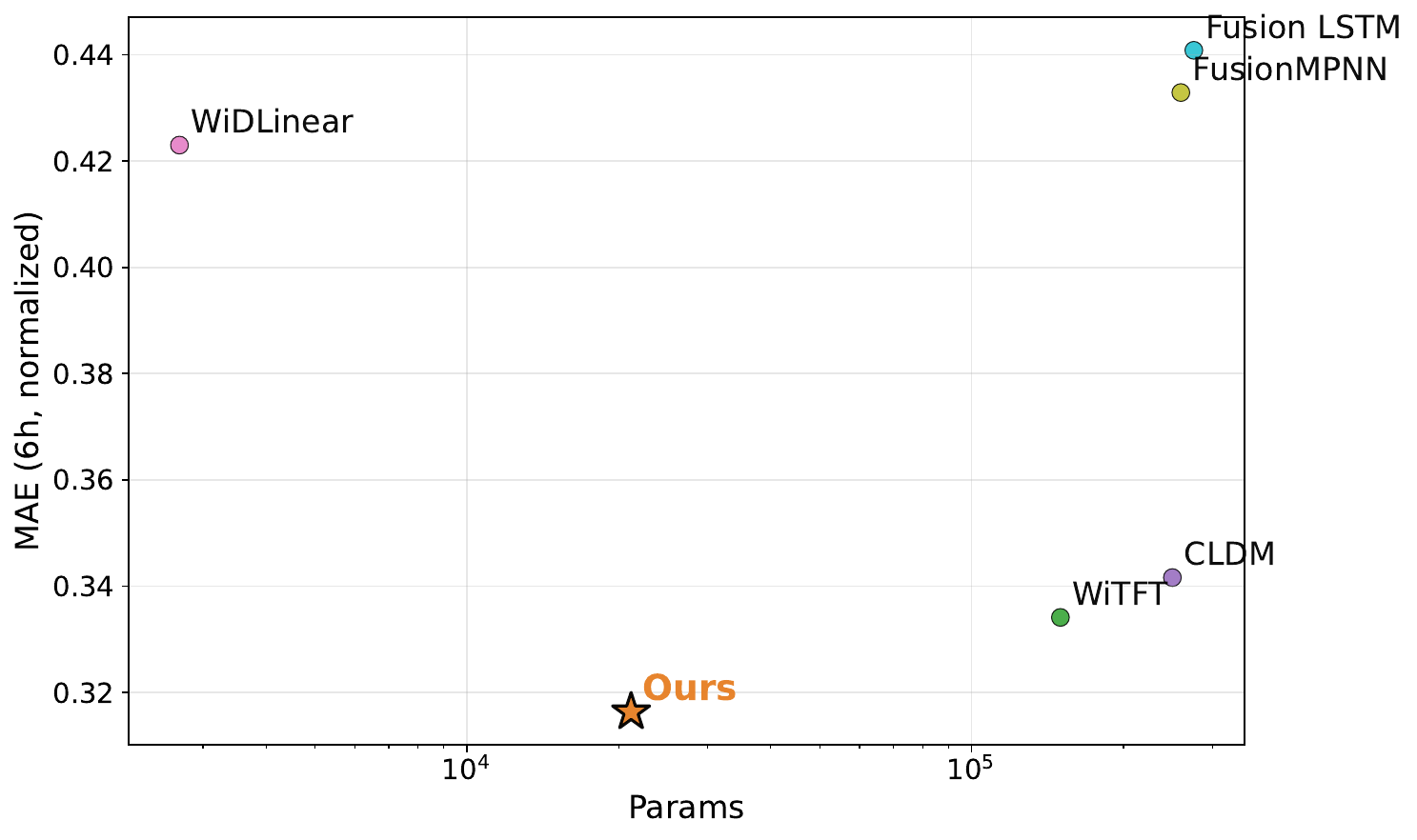}
    \caption{Parameter-efficiency analysis across different models. Our method achieves better performance with fewer parameters.}
    \label{fig:Scalability Analysis}
    \vspace{-10pt}
\end{figure}
As shown in the \cref{fig:Scalability Analysis}, our method achieves a competitive performance with significantly fewer parameters than more complex models such as CLDM and WiTFT. While these larger architectures benefit from richer modeling capacity—especially through attention or graph-based structures—our lightweight design demonstrates that effective feature fusion combined with geometric-aware encoding enables efficient learning of spatiotemporal patterns in NWP data. In contrast, WiDLinear, despite its minimal parameter count, suffers from limited expressiveness due to its simple linear structure. The high-performing but heavy models in the top-right (Fusion LSTM and FusionMPNN) suggest that mere temporal embedding aggregation without structural reasoning does not yield substantial gains. Our approach strikes a favorable balance between model complexity and performance, highlighting the importance of geometrically informed representations in achieving strong results with reduced computational cost.

\section{Conclusion}
In this work, we propose a novel geometric neural operator framework for short-term wind power forecasting, which effectively integrates SCADA data with CERRA numerical weather predictions. By introducing a geometric encoder that captures rotation-invariant features (such as wind-turbine alignment and large-scale flow descriptors) and employing Fourier neural operators to model long-range spatiotemporal dependencies, the proposed model overcomes the limitations of existing methods that inadequately exploit directional and atmospheric dynamics. Experimental results across three wind farms demonstrate that our approach consistently outperforms state-of-the-art baselines. Ablations further confirm the role of each component.

\section{Limitations and Ethical Considerations}
All data used in this study, including the wind farm data and CERRA reanalysis data, are open and public datasets from authoritative meteorological institutions, with data acquisition and usage fully complying with the relevant copyright and open access norms. No private, sensitive, or personal information is involved in the research. For limitation, current framework primarily leverages wind-related variables, and is only evaluated for forecast horizons up to 6 hours. In the future, we plan to incorporate more weather variables, such as temperature and geopotential, and extend it to forecasting with longer lead times.

\begin{acks}
This work is funded by National Natural Science Foundation of China Grant No. 72371217, the Guangzhou Industrial Informatics and Intelligence Key Laboratory No. 2024A03J0628, the Shenzhen Science and Technology Program (KCXFZ20240903093759004), the Nansha Key Area Science and Technology Project No. 2023ZD003, and Project No. 2021JC02X191.
Yang Liu is supported in part by the Postdoctoral Fellowship Scheme of The Chinese University of Hong Kong.
\end{acks}

\section{GenAI Disclosure}
In this work, we used LLMs as an auxiliary tool for polishing the writing and improving the clarity of the manuscript. Specifically, we use it to review the entire text for grammatical accuracy, improve sentence structure, and ensure consistent phrasing and tone throughout the paper. 

\newpage

\bibliographystyle{ACM-Reference-Format}
\bibliography{references}

\section{Appendices}
\setlength{\floatsep}{3pt plus 1pt minus 1pt}
\setlength{\textfloatsep}{3pt plus 1pt minus 1pt}
\setlength{\intextsep}{3pt plus 1pt minus 1pt}
\setlength{\abovecaptionskip}{3pt}
\setlength{\belowcaptionskip}{2pt}
\renewcommand{\arraystretch}{1}
\subsection{Dataset Construction}

\begin{table}[H]
\small
\centering
\caption{Dataset summary and forecasting setup.}
\label{tab:dataset_summary}
\resizebox{0.94\linewidth}{!}{%
\begin{tabular}{llcccc}
\toprule
\textbf{Farm} & \textbf{Period} & \textbf{Turb.} & \textbf{Grid} & \textbf{Input} & \textbf{Pred.} \\
\midrule
Kelmarsh      & Jan. 2016--Dec. 2020 & 6  & 16 & 9 & 6 \\
Penmanshiel   & Jun. 2016--Dec. 2020 & 14 & 20 & 9 & 6 \\
Hill of Towie & Jan. 2016--Dec. 2020 & 21 & 21 & 9 & 6 \\
\bottomrule
\end{tabular}
}
\medskip
\parbox{0.94\linewidth}{\raggedright\footnotesize Note: Turb. = turbines; Grid = matched CERRA points; Input/Pred. = 3h history/1h horizon. Sources: Kelmarsh~\cite{zenodo_kelmarsh_2022}, Penmanshiel~\cite{zenodo_penmanshiel_2022}, and Hill of Towie~\cite{zenodo_hilloftowie_2025}.}
\end{table}

We align SCADA to hourly timestamps and sample every 3 hours to form history features: power $P_3$, increment $\Delta P_3$, wind speed $W_3$, and direction/yaw sine--cosine terms.
CERRA inputs use the $K{=}16/20/21$ nearest grid points per farm: analysis fields aligned to the 3-hour history and forecasts issued at $t_0$ for lead times 1--6 hours, both with speed, direction, $u$, and $v$.
For each turbine--grid pair we also store normalized $(\Delta x,\Delta y,d)$ geometry.
We split chronologically by target time (train 2016--2018, validation 2019, test 2020), assigning a sample by its last forecasting hour to avoid temporal leakage.
All variables are z-score normalized with training-only statistics; invalid entries are set to zero after normalization, and CERRA forecasts use separate per-(lead, grid, variable) statistics.

\subsection{Implementation Details}\label{sec:imp}

\paragraph{\textbf{Our official implementation.}}
Our final model follows Fig.~\ref{fig:framework}: a non-parametric geometric encoder builds alignment/global-wind features, modality-specific MLPs project SCADA and CERRA into a shared latent space, and two frequency-fusion layers mix spectra along time.
We use $d_{\text{time}}{=}16$ and $d_{\text{hidden}}{=}8$, producing $\bm{h}_{s}^{(0)}\in\mathbb{R}^{16\times N\times 8}$ and $\bm{h}_{g}^{(0)}\in\mathbb{R}^{16\times K\times 8}$; $\bm{H}^{(2)}$ is flattened and decoded by a two-layer MLP.

\paragraph{\textbf{Optimization and training protocol.}}
We train with masked MSE and AdamW using learning rate $1\times 10^{-4}$, weight decay $1\times 10^{-4}$, batch size $128$, dropout $0.15$, and early stopping patience $20$ within $100$ epochs (\cref{tab:exp-setting-training}); no gradient clipping or learning-rate scheduling is used.
\paragraph{Hardware and software.}
All experiments are conducted using PyTorch [1.12] on GTX A6000, with CUDA [12.4].

\subsection{Baseline Implementation Details}

We summarize each baseline family; all learned models use the same split, seed where applicable, and comparable capacity for fair comparison.

\begin{table}[H]
\small
\centering
\caption{Common training settings.}
\label{tab:exp-setting-training}
\begin{tabular*}{0.94\linewidth}{@{\extracolsep{\fill}}lcc@{}}
\toprule
\textbf{Setting} & \textbf{ST Model / NWPFusion} & \textbf{TS Model} \\
\midrule
Batch size & 128 & 128 \\
Learning rate & $1\times10^{-4}$ & $1\times10^{-4}$ \\
Weight decay & $1\times10^{-4}$ & default \\
Max epochs & 100 & 100 \\
Early stopping & 20 & 20 \\
Dropout & 0.15 & 0.15 \\
\bottomrule
\end{tabular*}
\end{table}

\paragraph{\textbf{STModel (SCADA-only spatial--temporal).}}
Models in this family use only SCADA histories $(B,L,N,F)$ to predict $(B,H,N)$; graph variants also receive the turbine adjacency matrix, and patch variants use patch length 3.
We keep $d_{\mathrm{model}}{=}128$ and dropout 0.15.

\paragraph{\textbf{FNP baseline.}}
Our FNP baseline uses the same SCADA and CERRA task interface as the proposed model: raw/derived features are projected to $D{=}128$, temporally aligned, fused by one 8-mode neural Fourier layer, flattened, and decoded by an MLP to $(B,H,N)$.
No recurrent or graph decoder is used in this FNP implementation.

\paragraph{\textbf{Other SCADA+CERRA fusion baselines.}}
These baselines share an embedding stage: a shared GRU encodes SCADA histories, an MLP embeds matched CERRA forecasts flattened per turbine, and the two embeddings feed AR/NAR heads including LSTM, MLP, MPNN, AGCRN, DLinear, iTransformer, D2STGNN, MTGNN, HiSTGNN, and HimNet.
We use hidden dimension 128 and the same split, masks, and training protocol.

\paragraph{\textbf{Time-series and NWP-only baselines.}}
For TSLib models~\cite{wu2023timesnet}, SCADA is reshaped from $(T,N,F)$ to $(T,NF)$ and predicted as $(B,H,N)$; DLinear, TimesNet, TimeXer, iTransformer, TimeMixer, PatchTST, TSMixer, and WPMixer use $d_{\mathrm{model}}{=}d_{\mathrm{ff}}{=}128$.
WiTFT, WiDLinear, Basic Diffusion, and CLDM are NWP-only baselines using CERRA inputs $(B,7,64)$.

\subsection{Evaluation Metrics}
\label{sec:appendix-evaluation-metrics}

We report masked, node-aggregated MAE and RMSE over lead steps $\mathcal{H}_3{=}\{1,2,3\}$ and $\mathcal{H}_6{=}\{1,\ldots,6\}$. For a horizon set $\mathcal{H}$, target $y$, prediction $\hat{y}$, and validity mask $m$, the metrics are:
\begin{align*}
\text{MAE}(\mathcal{H}) &=
\frac{\sum_{i,t,h\in\mathcal{H}} m_{t,h}^{(i)} |y_{t,h}^{(i)}-\hat{y}_{t,h}^{(i)}|}
{\sum_{i,t,h\in\mathcal{H}} m_{t,h}^{(i)}},\\
\text{RMSE}(\mathcal{H}) &=
\sqrt{
\frac{\sum_{i,t,h\in\mathcal{H}} m_{t,h}^{(i)} (y_{t,h}^{(i)}-\hat{y}_{t,h}^{(i)})^2}
{\sum_{i,t,h\in\mathcal{H}} m_{t,h}^{(i)}}}.
\end{align*}
Only valid entries are included, and errors are aggregated jointly across turbines and time steps.

\begin{craddition}
\subsection{Additional Analyses}

\paragraph{\textbf{Operational error conversion.}}
\cref{tab:cr-kw-scale} reports the approximate conversion from normalized MAE to kW using the farm-specific scaling factors in the processed data.
A 0.01 change in normalized MAE corresponds to about 5.8--7.2 kW for each turbine-step, so the 0.005--0.016 gaps in the controlled experiments below remain operationally meaningful after aggregation over turbines, horizons, and repeated dispatch windows.

\begin{table}[H]
\centering
\caption{Approximate conversion from normalized MAE to kW.}
\label{tab:cr-kw-scale}
\resizebox{0.94\linewidth}{!}{%
\begin{tabular}{lccc}
\toprule
\textbf{Farm} & \textbf{$<3$h nMAE / kW} & \textbf{$<6$h nMAE / kW} & \textbf{+0.01 nMAE} \\
\midrule
Kelmarsh & 0.300 / 175.6 & 0.311 / 182.0 & $\sim$5.8 kW \\
Penmanshiel & 0.289 / 193.5 & 0.300 / 201.4 & $\sim$6.7 kW \\
Hill of Towie & 0.326 / 235.2 & 0.338 / 244.4 & $\sim$7.2 kW \\
\bottomrule
\end{tabular}
}
\end{table}

\paragraph{\textbf{Supplementary experiment settings.}}
UK-farm supplementary experiments reuse the main preprocessing, split, masks, and training protocol; controlled comparisons change only the tested component.
Tables~\ref{tab:cr-fusion-variants} and~\ref{tab:cr-ts-st-backbones} report normalized test MAE (mean $\pm$ std) over Kelmarsh, Penmanshiel, and Hill of Towie, while \cref{tab:cr-cross-site} reports two external sites separately.
The fusion variants keep the SCADA and CERRA inputs and decoder fixed, the matched-input backbones keep the shared SCADA+CERRA embedding protocol fixed, and the cross-site experiment changes the farm family, turbine layout, and data period.

\begin{table}[H]
\centering
\caption{Fusion variants: mean $\pm$ std MAE over the three UK farms.}
\label{tab:cr-fusion-variants}
\begin{tabular*}{0.94\linewidth}{@{\extracolsep{\fill}}lcc@{}}
\toprule
\textbf{Fusion} & \textbf{$<3$h MAE} & \textbf{$<6$h MAE} \\
\midrule
Concat & $0.330 \pm 0.016$ & $0.332 \pm 0.015$ \\
Conv & $0.330 \pm 0.016$ & $0.332 \pm 0.015$ \\
Pool-inject & $0.320 \pm 0.018$ & $0.324 \pm 0.016$ \\
Gate & $0.318 \pm 0.021$ & $0.326 \pm 0.018$ \\
FiLM & $0.312 \pm 0.019$ & $0.322 \pm 0.019$ \\
Bilinear & $0.312 \pm 0.019$ & $0.321 \pm 0.017$ \\
Cross-attn & $0.311 \pm 0.020$ & $0.323 \pm 0.018$ \\
Ours & $\mathbf{0.304 \pm 0.017}$ & $\mathbf{0.316 \pm 0.016}$ \\
\bottomrule
\end{tabular*}
\end{table}

Table~\ref{tab:cr-fusion-variants} isolates the fusion operator under fixed inputs and decoder.
Concat and Conv both give 0.332 at 6h, indicating that naive token merging or local temporal mixing is weak once point-level SCADA and gridded CERRA features are aligned.
Adaptive variants narrow the gap: FiLM, Bilinear, and Cross-attn reach 0.321--0.323 at 6h, but Ours remains best at 0.316.
The 0.005--0.007 6h margin over Bilinear/Cross-attn corresponds to several kW per turbine-step by \cref{tab:cr-kw-scale}, supporting the rebuttal point that the gain comes from spectral cross-modal alignment rather than only richer inputs.

\begin{table}[H]
\centering
\caption{Matched-input backbones: mean $\pm$ std MAE over the three UK farms.}
\label{tab:cr-ts-st-backbones}
\begin{tabular*}{0.94\linewidth}{@{\extracolsep{\fill}}lcc@{}}
\toprule
\textbf{Model} & \textbf{$<3$h MAE} & \textbf{$<6$h MAE} \\
\midrule
iTransformer & $0.312 \pm 0.011$ & $0.338 \pm 0.015$ \\
DLinear & $0.381 \pm 0.003$ & $0.421 \pm 0.004$ \\
MTGNN & $0.343 \pm 0.015$ & $0.388 \pm 0.013$ \\
HimNet & $0.330 \pm 0.009$ & $0.376 \pm 0.010$ \\
Ours & $\mathbf{0.304 \pm 0.017}$ & $\mathbf{0.316 \pm 0.016}$ \\
\bottomrule
\end{tabular*}
\end{table}

Table~\ref{tab:cr-ts-st-backbones} compares representative time-series and spatio-temporal backbones under the same SCADA and CERRA inputs, following the shared-input protocol in Xu et al.~\cite{xu2025benchmark}.
DLinear is weakest, iTransformer is competitive at 3h but degrades more at 6h, and MTGNN/HimNet add spatial bias yet remain 0.060/0.048 above Ours at 6h.
Because all methods receive matched weather signals, this gap reflects architectural design: our geometric encoder builds turbine--flow features before spectral fusion instead of leaving generic backbones to infer them from concatenated tokens.

\paragraph{\textbf{Cross-site generalization.}}
\begin{table}[H]
\centering
\caption{Cross-site MAE on external datasets.}
\label{tab:cr-cross-site}
\begin{tabular*}{\linewidth}{@{\extracolsep{\fill}}lcccc@{}}
\toprule
\textbf{Model} & \textbf{S $<3$h} & \textbf{S $<6$h} & \textbf{N $<3$h} & \textbf{N $<6$h} \\
\midrule
iTransformer & 0.226 & 0.257 & 0.271 & 0.343 \\
DLinear & 0.330 & 0.329 & 0.419 & 0.426 \\
MTGNN & 0.207 & 0.237 & 0.275 & 0.353 \\
HimNet & 0.276 & 0.286 & 0.269 & 0.340 \\
Ours & \textbf{0.204} & \textbf{0.229} & \textbf{0.257} & \textbf{0.317} \\
\bottomrule
\end{tabular*}
\end{table}

We further evaluate on SMARTEOLE (France, 2020)~\cite{duc2026smarteole} and Norre M2 (Denmark, 1991--1993)~\cite{norre_m2_2022}.
For fairness, baselines follow Xu et al.~\cite{xu2025benchmark} with the shared SCADA and CERRA encoder, whereas Ours keeps the submitted geoencoder and fusion pipeline under matched splits, masks, per-site normalization, horizons, and training budget.
Table~\ref{tab:cr-cross-site} broadens the evidence beyond the UK farms.
Ours improves over MTGNN on SMARTEOLE by 0.003/0.008 at 3h/6h and over the strongest Norre M2 baseline by 0.012/0.023.
The larger long-horizon margin on Norre M2 is consistent with the main experiments, where weather-aware geometric fusion becomes more useful as pure SCADA extrapolation weakens.
Because these sites differ in geography, turbine layout, and measurement coverage, this result is best viewed as a lightweight out-of-distribution check rather than another in-family validation.

\end{craddition}


\end{document}